# Addressing database variability in learning from medical data: an ensemble-based approach using convolutional neural networks and a case of study applied to automatic sleep scoring

Diego Alvarez-Estevez[1*], Isaac Fernández-Varela[2]

(1) Sleep Center, Haaglanden Medisch Centrum, Lijnbaan 32, 2512VA, The Hague, The Netherlands
(2) Laboratory for Research and Development of Artificial Intelligence, Computer Science Department, University of A Coruña, Campus Elviña s/n, 15071, A Coruña, Spain
* Corresponding author, email: diego.alvareze@udc.es

**Abstract**
In this work we examine some of the problems associated with the development of machine learning models with the objective to achieve robust generalization capabilities on common-task multiple-database scenarios. Referred to as the "database variability problem", we focus on a specific medical domain (sleep staging in sleep medicine) to show the non-triviality of translating the estimated model's local generalization capabilities into independent external databases. We analyze some of the scalability problems when multiple-database data are used as inputs to train a single learning model. Then, we introduce a novel approach based on an ensemble of local models, and we show its advantages in terms of inter-database generalization performance and data scalability. In addition, we analyze different model configurations and data pre-processing techniques to determine their effects on the overall generalization performance. For this purpose, we carry out experimentation that involves several sleep databases and evaluates different machine learning models based on convolutional neural networks.

## 1. Introduction

The continuous increase in data availability during the last years and recent advances in Machine Learning (ML), especially those associated with the outbreak of deep neural networks, have opened new possibilities for automatic pattern recognition in a number of domains. In this work we focus on the intelligent analysis of medical data.

In the context of ML, data from a single dataset will typically be presented to a learning algorithm. These data usually comprise examples of a task to be learned. Normally, part of the available data would be intentionally put aside, keeping it independent of the subset of data used during the training process. This subset of the data is commonly referred to as the "testing set". Testing data are used to estimate the generalization capabilities of the model that simulates the conditions in the final operational environment, when the model is presented with new (unseen) examples. Often, however, when examining generalization within a broader context, achieving good performance on the testing set does not guarantee the capacity of the model to fully abstract the real underlying task from the specific data used for its development. In fact, when considering two or more datasets from independent sources, even if referring to the same common task, data will likely present differences due to the specifics of the respective dataset generation procedures. In the medical domain, for example, the same physiological variable might have been monitored concurrently by two different acquisition systems, each characterized by a different signal-to-noise ratio, and thus generating different data. Consequently, a learning model that was only presented with examples from one of the two systems might have trouble when handling data from the other. Effectively, even though the performance of a ML model might have been "independently" evaluated using its corresponding local testing set, in practice, few (if anything at all) can be concluded with regards to the expected performance of the model when presented with data from an independent external dataset. Surprisingly, little attention has been paid in the related scientific literature to assess the so-called "database variability problem". For this purpose, generalization capabilities of a ML model should be evaluated in a broader scenario, namely, by considering two or more independent data sources in reference to a common task in which data have been kept independent of the model's development and parameterization process.

Here, we focus on this topic, which we will illustrate with an example from a specific medical domain. In particular, and without loss of generalization, we will discuss the case of automatic sleep staging in the field of sleep medicine. Sleep staging characterizes the patient's sleep macrostructure, and is one of the most important tasks within the context of sleep medicine. This characterization results in the so-called hypnogram, which summarizes the evolution of the different sleep states from the voluminous chart recordings of electrical activities recorded throughout the night. These recordings, which are referred to as polysomnographic (PSG) tests, are the gold standard reference for the diagnosis of many sleep disorders.

The first official guidelines for the analysis of sleep and construction of the hyponogram were initially proposed by Rechtschaffen and Kales (R&K) in 1968 [1]. They proposed the assignment of six possible sleep state labels, namely, wakefulness (W), stages 1–4 (S1, S2, S3 and S4), and rapid eye movement (REM) to characterize each discrete time epoch interval of 30s. Epoch classification into each of the sleep stages is performed according to the observed pattern of neurophysiological activity in the corresponding PSG interval. This activity involves the monitoring of different electroencephalographic (EEG), electromyographic (EMG) and electrooculographic (EOG) signal derivations. Since 2007 this standard has been periodically reviewed by the American Academy of Sleep Medicine (AASM) [2] [3]. The AASM hypnogram is also a 30s epoch-based segmentation of the sleep process but, among other differences, the four non-REM stages (S1-S4) were reduced to three (N1, N2 and N3).

Automatic analysis of sleep macrostructure is of interest because of the complexity and high costs associated with human scoring. In fact, the topic dates back to the 1970's when the first approximations came into sight. Countless attempts have been appearing since then (see [4] [5] [6] [7] [8] [9] [10] [11] [12] [13]) evidencing that the task still represents an open challenge, and an active area of research interest. More recently, several publications have emerged that are based on the use of deep learning, and claim advantages over previous approaches that include improved performance, and the possibility to skip handcrafted feature engineering processes [14] [15] [16] [17] [18]. A common drawback, however, has to do with the limited validation procedures presented in these studies with regard to the aforementioned database variability problem. First, most of the approaches reported so far have been validated using just one individual dataset. Moreover, this dataset is usually rather small, controlled, and many times, not publicly accessible, which prevents repeatability of the experiments. In the few exceptions where more than one dataset was approached, the validation was either carried out by re-parameterizing of the model for each dataset, or no independent database-wise separation was performed between the training and the testing data. As stated before, this fact limits the ability to make predictions about the actual generalization capabilities of these algorithms when confronting larger, independent, and heterogeneous databases. The actual reality, in fact, is that the grade of acceptability for these algorithms among the clinical community remains low, being rarely used in the clinical practice. To our knowledge only a few works [19] [20] [21] [22] have reported generalization results using independent external databases.

In the next sections we further develop these concepts. First we particularize the database variability problem in the context of the sleep scoring task. Then, from a general point of view, we analyze some of the problems in the design of ML models to achieve generalization capabilities among different datasets. After that, we introduce a novel approach based on an ensemble of individual learning models that solves some of the problems associated with such a design. To test our hypotheses we carry out different experiments, showing how the proposed approach can effectively improve the generalization capabilities over the individual models derived from each dataset. For this purpose, we use as a reference different PSG source datasets that contain large collections of physiological data, training and evaluating different ML models using Convolutional Neural Networks (CNNs). After analyzing the results, we finish discussing the implications of our findings and enumerate the main conclusions of this work.

## 2. The database variability problem in the context of sleep studies

While clinical standard guidelines, such as those contained in the R&K or AASM manuals, aim for a certain level of homogenization, in practice different sources of uncertainty and variability affect the recording and the analysis of PSG data. It is well-known, for example, that differences regarding subjects' conditions and related physiology have influence on the resulting sleep micro and macrostructure signatures. For instance, sleep-disordered patients tend to present a disrupted sleep cycle with frequent sleep-wake transitions and reduced periods of slow wave sleep, as with respect to healthy sleepers [23]. Age is also known to be a modulating factor associated with larger proportions of REM and deep sleep in infants and children, in comparison to elderly adults. In the second group, in contrast, sleep is lighter and more fragmented with higher presence of brief EEG arousals or longer awakenings throughout the night [24]. Differences in electrode positions can also introduce additional sources of data variability, as the expression of different sleep microstructure events is clearly localized among specific brain regions. K-complexes or vertex-waves, for example, are usually more visible across the frontal lobe areas [25]. Similarly, the digital sampling frequency determines the range of valid frequencies for signal analysis, and therefore controls the resolution and quality of the recorded sleep pattern activity.

In general, sources of variability can be very diverse, making PSG scoring a particularly challenging field. A comprehensive explanation of all the influencing factors, and their consequences on the analysis, would require extensive discussion, which falls beyond the scope of this work. For reference, Table 1 summarizes some of the most important aspects, together with some specific examples on the sleep staging task. For more details, the interested reader is aimed to consult the classical literature on clinical neurophysiology and digital signal acquisition.

Table 1. Different sources of variability affecting the recording and interpretation of physiological signals. For the examples we will assume A and B to be two independent data sources containing data with regard to the sleep staging task

| Category | Subcategory (Differences in...) | Examples | Example in sleep monitoring |
|---|---|---|---|
| Variability in the subject's source population | Subject's conditions | Diseased or healthy patients, differences in age, use of medication, the selection criteria, etc. | Dataset A: Young healthy subjects with no medication<br>Dataset B: Old insomnia patients using sleep medication |
| | Subject's physiology | Skull thickness, gender, anatomical differences in general | Dataset A: Obese population with equal gender distribution<br>Dataset B: Population composed only of female subjects |
| | Class distributions (classification tasks) | Different proportion of output targets | Dataset A: 20% of scored stages belong to REM sleep<br>Dataset B: 5% of scored stages belong to REM sleep |
| Variability regarding data recording and/or acquisition | Monitoring/acquisition | Different sensors/transducers | Dataset A: Airflow recording with nasal cannula<br>Dataset B: Airflow recording with thermistor |
| | | Electrode positions, impedances | Dataset A: EEG recorded at Cz location with 5 KΩ resistance<br>Dataset B: EEG recorded at Fpz location with 1 KΩ resistance |
| | | Amplification factors, sampling rates, pre-filter settings | Dataset A: EEG recorded at 125 Hz with high-pass filter at 10 Hz<br>Dataset B: EEG recorded at 512 Hz with low-pass filter at 100 Hz |
| | | Noise, interference, and monitoring artifacts | Dataset A: Clean EEG recording<br>Dataset B: EEG with constant presence of muscle and ECG artifact |
| | | Bit resolution, digital storage format | Dataset A: Data recorded using 24 bits per sample<br>Dataset B: Data recorded using 8 bits per sample |
| Variability due to data interpretation | Reference scoring standards | E.g. for sleep scoring: AASM (and their different versions) vs R&K | Dataset A: Sleep scored according to AASM 2012 guidelines<br>Dataset B: Sleep scored according to R&K guidelines |
| | Intra-scorer subjectivity | Tiredness, stress, general mood... | Dataset A: Scoring has taken place in good mood and after restful sleep<br>Dataset B: Sleep scorer is stressed due to huge working load and had very bad sleep the day before |
| | Inter-scorer subjectivity | Subjective interpretation of the same phenomena by two different persons, different training and/or experience, use of different ad-hoc rules | Dataset A: Data scored by young technician trained at sleep center A<br>Dataset B: Data scored by experienced technician who had experience in different sleep centers |

For our purpose, it is important to bear in mind is that all these sources of variability trigger divergences in the interpretation by different human experts [26] [10] [27] [28] [29] [30] [19] [31]. Similarly, when an algorithm learns from a specific sleep dataset, it will be influenced by the specific characteristics of its data. Furthermore, a learning algorithm will be sensitive to additional sources of variability, usually not affecting the human expert, derived from the structure of the data, and the specifics of machine learning methods. For example, a ML algorithm might be conditioned by the presence of data outliers or imbalanced class distributions on the training dataset [32]. Inter-database variability hence represents a challenge that makes it difficult for an algorithm to generalize its performance from one database to another.

Specific experiments using real patient data will be carried later in this work, to illustrate and quantify this problem on the context of sleep scoring. Meanwhile, in the following section, discussion will be carried out with regard to the difficulties on the design of scalable ML models, in particular, when confronting different heterogeneous databases with the objective to achieve good inter-database generalization performance.

## 3. Scalability problems when learning from multiple datasets

The most intuitive approach to encourage good generalization performance of a ML model is to use as much input data as possible during the learning process. In the scenario where data from different sources are involved, the former would translate in using data from the all the available datasets. Thereby the amount of training data increases, as well as their heterogeneity, hence boosting the chances of coming up with a true generalist model by minimizing the dataset overfitting risk. This approach, however, has its own drawbacks.

When training a unique model, the first and most obvious consequence is that, by increasing the amount of input data, more computational cost will also be needed for the training process. The extra cost does not only impact the execution time, but as the number of involved datasets increases, memory resources need to be increased as well to keep data available during training. In addition, one should take into account that training a ML model usually involves several repetitions due to hyperparameter optimization. Considering that just one repetition can be already very costly in Deep Learning, depending on the size of the dataset and the available resources, the required increment in computational costs might be ultimately unattainable.

On top of that, an additional inconvenient of this approach concerns the lack of flexibility as the number of available datasets dynamically evolves in time. Indeed, suppose one initially has 3 datasets available, *A*, *B*, and *C*, for training a model *M(ABC)*. If later on, a new dataset becomes available, namely *D*, there is no option but to train a whole new model *M(ABCD)*, using *A*, *B*, *C*, and *D* as input. Otherwise, retraining *M(ABC)* using *D* as the only input would degrade the past learning (i.e. the model will "forget" about A, B and C data). In ML literature, this effect is termed *catastrophic forgetting* [33].

Finally, one closely related, but perhaps less obvious problem, has to do with the difficulty of exploring all the different input dataset combinations with the objective to find the best possible model in terms of the resulting inter-database generalization capabilities. To illustrate that, let us assume a classical data partitioning schema [32] in which data from dataset *X* are split into $X_{TR}$, $X_{VAL}$, and $X_{TS}$ subsets, respectively, for the purposes of training (*TR*), validation (*VAL*), and testing (*TS*) of a model. Notice that, in the case of neural networks, the validation subset can be used as well to implement the early stopping criterion [32]. Leaving aside the specific proportion of *X* data assigned to each of the *TR*, *VAL*, and *TS* subsets, let us also assume that *N* independent datasets are available as input, and that all the corresponding $X_{TS}$ subsets are set aside as independent testing data. Under these circumstances, it can be shown (see Supplementary materials) that the number possibilities to combine the different $X_{TR}$, $X_{VAL}$ subsets (*2N* subsets in total) into individual (*TR*, *VAL*) pairs, has exponential complexity $O(2^{2N})$. Notice that although some of the combinations might represent more logical choices a priori (as stated before, for example, intuition might lead the data scientist to explore the combinations involving the most input data), in general it is not possible to know beforehand which combination will lead to the best model in terms of inter-database generalization. Thus, in practice, one should try all the combinations systematically to be sure. Once again, each combination involves running a separated training process with its corresponding repetitions for hyperparameter optimization. Normally, the computational costs associated with this approach are unacceptable.

## 4. An alternative approach: ensemble combination of local models

We propose an approach to cope with the scalability problems described in the previous Section 3. Under this approach we train one independent model for each dataset available, and then combine the resulting individual models using an ensemble method. Specifically, in this work we are assuming the ensemble output takes place using the majority vote [34] [35].

Thus, let us suppose three datasets are available for training purposes (*A,B,C*) as in the previous section. Each dataset then leads to one local independent model: *M(A)*, *M(B)*, and *M(C)*. Given any hypothetical testing dataset, prediction takes place using the ensemble *ENS[M(A), M(B), M(C)]*. Obviously, this has the advantage that no big *TR* dataset needs to be collected or be kept available "as a whole" in memory, given to the fact that each dataset *X* is local to the corresponding individual model *M(X)*. Moreover, when a new dataset *D* becomes available, then it is enough to train the new local model *M(D)*, and eventually add it to the ensemble: *ENS[M(A), M(B), M(C), M(D)]*, without the need to modify any past learning.

Finally, notice that for each dataset *X* leading to the individual model *M(X)*, and following the same schema as in Section 3, the (*TR*, *VAL*) partitioning combinations are limited to either ($X_{TR}$, -) or ($X_{TR}$, $X_{VAL}$). Under these circumstances, it can be shown that the resulting combinatorial search space still has exponential complexity. However, the effective training space significantly reduces from exponential to *2N* (linear) complexity (see Supplementary materials for more details). Indeed, as under the majority vote schema only the votes of the individual models need to be recomputed, there is no need to retrain any model, no matter the combination to be tested in the ensemble. Furthermore, the computational cost associated with the training of each of these individual models is also much more reduced, again, as under the proposed ensemble schema the training of $M(X)$ does only involve data from the local dataset $X$.

## 5. Experimental methods

To quantify the database variability problem, and in order to test the proposed approach, the following experimentation was carried out:

Different clinical sleep scoring datasets, each one from one independent database source, were collected. A description of the characteristics of each dataset is provided in the next subsection 5.1. For the purposes of reproducibility, all databases were gathered from public online repositories, with the only exception of our own local sleep center database (which at present cannot be made publicly available due to patient privacy reasons). With no exception, all the databases are digitally encoded using the open EDF(+) format [36] [37].

For each dataset $k$, $k=1...K$, a ML model $M(k)$ was trained using a Convolutional Neural Network (CNN) architecture. Several architecture variations and training procedures were investigated, and compared, in order to analyze their impact on the inter-database generalization capabilities of the resulting models.

For each CNN architecture variant, and for each dataset $k$, the following experiments were carried out:

Experiment 1:
-Each dataset $k$ is partitioned as $TR(k)$, $VAL(k)$ and $TS(k)$, respectively as training, validation, and testing data. Let us denote the whole dataset $k$ by $W(k)$. A model $M(k)$ is derived by learning from $TR(k)$ and using $VAL(k)$ for hyper-parameterization purposes. The "local" generalization performance of the resulting model $M(k)$ is measured by predicting the corresponding $TS(k)$ data. This is the performance that is usually reported in the literature when data from only one database are used for experimentation.

Experiment 2:
-Each resulting model $M(k)$, is used to predict the reference scorings on each of the complete datasets $W(j)$, $j=1...K$. By comparing the results of *Experiment 1* and *Experiment 2* one can assess the effects of the database variability problem. In effect, for each $M(k)$, one can compare the expected local generalization achieved in $TS(k)$, with the effective "inter"-database generalization performance on each W($j$), $j<>k$. Notice that when $j=k$ the results are biased since $TR(k) => M(k)$ and $TR(k) \subseteq W(k)$.

Experiment 3:
-Each dataset $W(k)$ is predicted by an ensemble $ENS(k)$ of the individual models $M(j)$, $j=1...K$, $j<>k$. E.g. $ENS(2) = ENS[M(1), M(3),..., M(K)]$. That is, excluding the $M(k)$ trained with the data in $k$, thus keeping $W(k)$ completely independent of $ENS(k)$. By comparing the results of *Experiment 3* with those of *Experiment 1* and *Experiment 2*, it is possible to assess the effects of the proposed ensemble approach in terms of the resulting inter-database generalization.

For evaluating the performance on each of the experiments the Cohen's kappa index ($\kappa$) was used as the reference validation metric. Cohen's kappa is preferred over other common validation indices (e.g. classification error, sensitivity, specificity, or $F_1$-score) as it accounts for the agreement due to chance, showing robustness in the presence of various class distributions, and thus allowing performance comparison among the different datasets [38]. Noteworthy, Cohen´s kappa is also the most widespread validation metric among studies analyzing human inter-rater variability in the context of sleep scoring.

### 5.1. Datasets

An overview of the different PSG datasets used in our experimentation is given in the following lines. Datasets were collected from different heterogeneous and independent database sources. An extended description of the individual datasets and the corresponding signal montages can be found in Supplementary Table 1.

Haaglanden Medisch Centrum Sleep Center Database (HMC)
This dataset includes a total of 159 recordings gathered from the sleep center database of the Haaglanden Medisch Centrum (The Netherlands) during April 2018. Patient recordings were randomly selected and include a heterogeneous population which was referred for PSG examination on the context of different sleep disorders. The recordings were acquired in the course of common clinical practice, and thus did not subject people to any other treatment nor prescribed any additional behavior outside of the usual clinical procedures. Data were anonymized avoiding any possibility of individual patient identification. This is the only dataset which is not publicly available online.

St. Vicent's Hospital / University College Dublin Sleep Apnea Database (Dublin)
This database contains 25 full overnight polysomnograms from adult subjects with suspected sleep-disordered breathing. Subjects were randomly selected over a 6-month period (September 02 to February 03) from patients referred to the Sleep Disorders Clinic at St Vincent's University Hospital, Dublin, for possible diagnosis of obstructive sleep apnea, central sleep apnea, or primary snoring. This database is available online through the PhysioNet website [39].

Sleep Health Heart Study (SHHS)
The Sleep Heart Health Study (SHHS) is a multi-center cohort study implemented by the National Heart Lung & Blood Institute to determine the cardiovascular and other consequences of sleep-disordered breathing. This database is available online upon permission at the National Sleep Research Resource (NSRR) [40] [41]. More information about the rationale, design, and protocol of the SHHS study can be found in the dedicated NSRR section [41], and in the related literature [42] [43]. For this study a random subset of 100 PSG recordings were selected from the SHHS-2 study.

Sleep Telemetry Study (Telemetry)
This dataset contains 44 whole-night PSGs obtained in a 1994 study of temazepam effects on sleep in 22 caucasian males and females without other medication. Subjects had mild difficulty falling asleep but were otherwise healthy. The PSGs were recorded in the hospital during two nights, one of which was after temazepam intake, and the other of which was after placebo intake. Subjects wore a miniature telemetry system which was described in the work of Kemp et al. [44]. Subjects and recordings are further described in another study [45]. The dataset is available online as part of the more extensive Sleep-EDF database at the PhysioNet website [46].

DREAMS Subject database (DREAMS)
The DREAMS dataset is composed of 20 whole-night PSG recordings coming from healthy subjects. The database was collected during the DREAMS project, to tune, train, and test automatic sleep stages algorithms [47]. The database is granted by University of MONS - TCTS Laboratory (Stéphanie Devuyst, Thierry Dutoit) and Université Libre de Bruxelles - CHU de Charleroi Sleep Laboratory (Myriam Kerkhofs) under terms of the Attribution-NonCommercial-NoDerivs 3.0 Unported (CC BY-NC-ND 3.0) [48].

ISRUC-SLEEP Dataset (ISRUC)
This dataset contains data from 100 adult subjects with evidence of having sleep disorders included in the ISRUC-Sleep database. Each recording was randomly selected between PSG recordings that were acquired at the Sleep Medicine Centre of the Hospital of Coimbra University (CHUC), in the period 2009–2013. More details about the rationale and the design of the database can be found in Khalighi et al. [49]. The database can be publicly accessed online [50].

For all the datasets, no exclusion criteria were applied a priori. Thus, all the recordings integrating the original selection were included on each simulation. The motivation was to assess the reliability of the algorithm on the most realistic situation, and to include the most general patient phenotype possible.

From each dataset two channels of EEG, the submental EMG, and one EOG derivation were extracted and used as input to the corresponding ML model. When available, one ECG derivation was used for the purposes of artifact filtering as optional pre-processing step (but not as input to the learning model, see subsequent Section 5.2). Recall from Table 1 that due to the specifics of each database, the effective signal montages differ per dataset. Supplementary Table 1 summarizes the main characteristics, and the specific selected derivations, according to the available montage on each case. In general, the followed rationale was to select two central EEG derivations, when possible, each one referencing to a different hemisphere (e.g. C4/M1 and C3/M2). If central derivations were not available, then frontal electrodes were chosen as backup. In some cases, no choice was possible according to this rationale, and therefore the only available derivations were used (e.g. for Telemetry, Pz-Oz and Fpz-Cz). In the case of the EOG, horizontal derivations were preferred as they are less sensitive to EEG and movement artifacts. We used the AASM scoring standard as reference for the output class labels. When the original dataset was scored using the R&K method, NREM stages 3 and 4 were merged into the corresponding N3, following the AASM guidelines.

## 5.2. Learning model

As stated in the previous sections, the learning model was implemented using Convolutional Neural Networks (CNNs). The general used architecture was based on a previous model developed by the authors [18].

Specifically, the CNN receives as input a 30 s window sample from each of the input signals (2 EEG channels, chin EMG, and horizontal EOG). As explained in Section 2, and among other differences, the sampling rate of each of the signals depends on the montage configuration of the source dataset. Thus, for a model to be able to process data from the different datasets, a common reference input needs to be set. For this purpose, we have opted to resample all signals at 100 Hz, representing a compromise between constraining the input dimensionality, and the preservation of the useful signal properties to carry out sleep scoring. Specifically, sampling at 100 Hz allows a working frequency up to 50 Hz which captures most of the interesting EEG, EMG and EOG frequencies. In this manner, each input to the network resulted on a matrix with size 4x3000.

The general network architecture is shown in Figure 1.

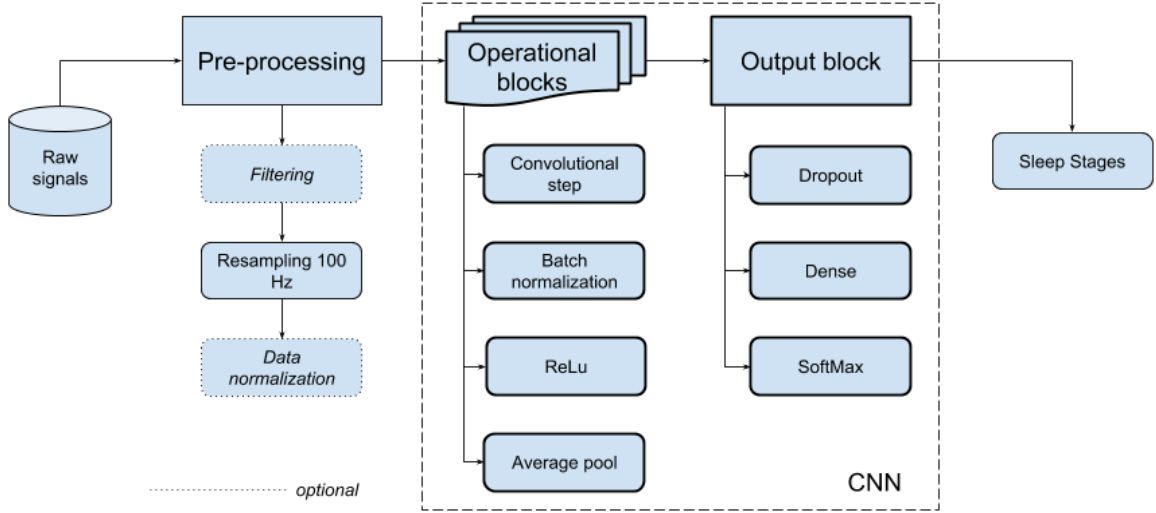

*Figure 1. Schema of the general CNN architecture used for training of the models*

The network architecture is composed of a concatenation of $N$ operational blocks. Each operational block $B(k)$, $k = 1...N$, is a set of four layers including a convolution step, that preserves the input size (with padding, dimension stride = 1), a batch normalization layer [51], a ReLu activation [52], and an averaging pool that reduces the input by a factor of two. The number of operational blocks $N$ is left as a hyperparameter to be configured. While all these operational blocks maintain the same kernel size, the number of filters in $B(k)$, was set as being two times the number of filters in $B(k-1)$. For $B(1)$, the initial number of filters was fixed to 8, based on previous experiments [18]. The specific type and the size of the convolutional kernel are also left as free parameters in this work.

After the last operational block, a dropout layer (0.5 probability) was added to improve regularization. Finally, a final dense full-connected layer with *softmax* activation was used at the output. The output of the *softmax* is interpreted as the posterior class probability for the corresponding input. Hence the node with the highest probability determines the classification decision.

For the learning algorithm the stochastic gradient descent was used. The maximum number of epochs was established to 30, and the initial learning rate was set to 0.001, decreasing by a factor of 10 every 10 epochs (thus $10^{-4}$, $10^{-5}$, up to a minimum of $10^{-6}$). For the data partition each dataset was split into $TR = 80\%$, $TS = 20\%$, with $VAL = 20\%$ of the $TR$ data. The training batch was set to 100 patterns, which was imposed by the available hardware resources and the size of the tested networks. As early stopping criterion we took as reference the validation loss, which was evaluated 5 times per training epoch, establishing a patience of 10, meaning that the training is stopped if the validation loss does not improve after two epochs (i.e. the whole training dataset was presented two times).

For each experiment the same random initialization seed was used to exclude variability due to random weights and data partition initialization, enabling deterministic training. This is important to assess influence of the different tested configurations, as described in the next subsections 5.2.1 and 5.2.2., and to make comparisons among the different models and datasets.

The cost function guiding the weight's update was a weighted version of the cross-entropy loss. We explored different options for the assignment of the class weights as a configuration parameter.

### 5.2.1. Model parameterization
The quest to achieve the maximum possible performance for a specific dataset is not the primary objective in this work. Instead, it is the focus to analyze the generalization performance of a relatively good general architecture on the multi-database prediction scenario. In this regard, exhaustive exploration of the hyper-parameter space was not a priority, and many parameterization decisions were based on previous experimentation described elsewhere [18].

However, we did found interesting to analyze the effect over the resulting inter-database generalization capabilities of some variants of the base model configuration. In particular:

-The cost function

Given the heterogeneous class distributions among the different datasets, we wanted to compare the assignment of proportional weight penalties in the calculation of the cross-entropy loss function to the results using the corresponding unweighted version.

When using the weighted cross-entropy loss, the weights $w(i)$ for each class $i$, $i=1...K$, were calculated according to the following formula:

$$w(i) = (1/card\,(i))/\sum_{j=1}^{K}\ \ 1/card(j)$$

where $card(i)$ denotes the cardinality, i.e. the number of elements in the class $i$. For the unweighted version all weights were set to $w(i) = 1$.

We hypothesize that the use of the weighted cross-entropy can result in better inter-database prediction capabilities of the resulting models. A reason for that is that the training of the model becomes less sensitive to the particular class distribution in the training dataset (i.e. learning would not be biased toward the majority classes).

-Convolution kernel size

We explored the use of different spatial representations on the feature space for the convolution step. For this purpose, we studied the use of different kernel configurations. Based on previous experimentation [18] we took as reference baseline a 1D convolution of size 1x10 (0.1 s) and then explored variations by *(i)* increasing the kernel in the time dimension up to 1x100 (1 s), and *(ii)* using 2D convolutions (kernel size of 4x10). In principle the use of 1D convolution would avoid imposing a fixed spatial structure (a priori unknown) between the different input signals.

-Depth of the network

The motivation here was to explore the relationship between the depth of the network and the generalization capabilities of the resulting models. Increasing the depth of the network can improve the generalization performance on the local dataset measured in *TS*. On the other hand, in the multiple-database scenario, there is still a risk of over-fitting the local dataset, translating into generalization decay when predicting external databases. In order to test the possible correlation between the local improvement and the inter-database generalization capabilities, we repeated all the experiments by doubling the initial number of operation layers from *N*=3 to *N*=6.

5.2.2. Data pre-processing

We also explored different data pre-processing methods to study their influence on the inter-database generalization capabilities of the resulting models. In particular, we explored the influence of different approaches for data normalization, and signal filtering for the removal of recording artifacts.

-Data normalization

Input data normalization is a common pre-processing step in ML. Especially in neural networks, data standardization is recommended to speed-up the training process, and to minimize the possibility to get trapped into local minima [32] [51].

On the multiple-database prediction scenario we hypothesize that non-normalized data can result in worst generalization capabilities, as digital ranges of the input signals are patient- and montage-dependent. For the same reason, however, data standardization based on long-term data trends, or whole-TR based normalization, can bias the learning process toward the characteristics of the reference dataset. The rationale here is that a "normalization factor" that standardizes the input distribution in one dataset can lead to a non-standard distribution (i.e. mean different from 0, and standard deviation different from 1) when applied to another dataset. Thus, some sort of "local" normalization might be more beneficial.

For testing this hypothesis we have repeated all the experiments by using *(i)* the non-normalized "raw" data as input, *(ii)* a dataset-based standardization using as reference the corresponding whole-TR data, and *(iii)* an epoch-based standardization by which each row of the 4x3000 input matrix is normalized to have zero mean and unit standard deviation.

-Signal filtering

Removal of artifacts and noise from the signals can eliminate "distracting" information hindering the generalization of the resulting models. Sources of artifacts might be coupled (at least partially) to the source datasets, hence inter-database generalization capabilities might benefit from this "data homogenization process".

Signal filtering was tested for general signal conditioning and artifact processing. When applied, this optional step was performed over the original raw signals, i.e. at the original signal sampling frequencies before resampling them to 100 Hz. For this purpose the following filtering pipeline was applied:

-Notch filtering: To remove the interference caused by the power grid. An AC frequency of 50 Hz is used in Europe, for example, while a frequency of 60 Hz is used in North America, therefore causing interference affecting signals at different frequency ranges depending on the source dataset. Design and implementation of the used digital filter has been described in previous works [53] [54].

-High-pass filter: Applied only to the chin EMG, it is meant to get rid of the DC and low frequency components which are not related to the (baseline) muscle activity. A first order implementation using a cut-off value at 15 Hz was used which has been described elsewhere [54].

-ECG filtering: Only when the ECG signal is available on the corresponding montage (no ECG derivation, for example, is available in the Telemetry dataset). An adaptive filtering algorithm was used in order to get rid of possible ECG artifact causing the appearance of spurious twitches on the input signals (EEG, EOG and EMG). The algorithm has been described in detail in Alvarez-Estevez et al. [53].

## 6. Results

In the following tables we show the results of the different experiments carried out as described in Section 5. For the sake of text economy, as well as for clarity, we are not including all the possible model configuration combinations. Instead, we focus the attention only on the most relevant results. At this purpose we are taking as basis the model parameterized with the weighted cross-entropy loss, a 1x10 convolution filter, 3-layer operational blocks, epoch-based input data normalization, and signal filters, then showing the resulting variations by changing the different configuration options as described in Sections 5.2.1 and 5.2.2. The model configuration combinations omitted from the tables do not change, by any means, the results of the analyses that follow in the subsequent sections.

Experiments were carried out on a laptop computer with Intel® Core™ i7-6700HQ CPU at 2.60 GHz, 16 GB of RAM, SSD storage access, and Nvidia GeForce GTX 960M GPU, using Matlab R2017b software. The computation time for the training of each individual model varied in the order of a few to several hours, as a function of the network configuration (e.g. number of filters, or depth of the network), the number of training patterns on the corresponding training dataset, and the number of training iterations.

Table 2 shows the results of the first experiment, where each of the learning models is trained and tested using data local to one dataset only.

*Table 2. Performance results of each individual model on the corresponding training (TR), validation (VAL) and testing (TS) partitions for each dataset. The number of effective training iterations is given in the third column. Results are reported as kappa index with respect to the corresponding reference clinical scorings.*

| Dataset | Model configuration | Training iterations | TR | VAL | TS |
|---------|--------------------|--------------------|------|------|------|
| HMC | Base model | 16 | 0.70 | 0.69 | 0.62 |
| | Non-weighted cross-entropy | 7 | 0.75 | 0.69 | 0.63 |
| | 1x100 convolutional kernel | 13 | 0.72 | 0.70 | 0.65 |
| | 4x10 convolutional kernel | 19 | 0.72 | 0.69 | 0.63 |
| | 6-layer operational blocks | 8 | 0.75 | 0.67 | 0.61 |
| | Without filtering | 15 | 0.70 | 0.67 | 0.61 |
| | TR-based data normalization | 15 | 0.70 | 0.70 | 0.62 |
| | Without data normalization | 12 | 0.69 | 0.68 | 0.61 |
| Dublin | Base model | 22 | 0.71 | 0.58 | 0.44 |
| | Non-weighted cross-entropy | 9 | 0.80 | 0.55 | 0.41 |
| | 1x100 convolutional kernel | 15 | 0.71 | 0.58 | 0.48 |
| | 4x10 convolutional kernel | 13 | 0.66 | 0.51 | 0.33 |
| | 6-layer operational blocks | 8 | 0.74 | 0.53 | 0.41 |
| | Without filtering | 14 | 0.72 | 0.60 | 0.46 |
| | TR-based data normalization | 13 | 0.62 | 0.55 | 0.45 |
| | Without data normalization | 12 | 0.68 | 0.53 | 0.43 |
| SHHS | Base model | 8 | 0.70 | 0.67 | 0.65 |
| | Non-weighted cross-entropy | 7 | 0.81 | 0.72 | 0.70 |
| | 1x100 convolutional kernel | 9 | 0.71 | 0.67 | 0.64 |
| | 4x10 convolutional kernel | 15 | 0.72 | 0.68 | 0.66 |
| | 6-layer operational blocks | 6 | 0.71 | 0.64 | 0.62 |
| | Without filtering | 14 | 0.72 | 0.68 | 0.64 |
| | TR-based data normalization | 12 | 0.75 | 0.69 | 0.69 |
| | Without data normalization | 19 | 0.74 | 0.68 | 0.67 |
| Telemetry | Base model | 12 | 0.73 | 0.65 | 0.58 |
| | Non-weighted cross-entropy | 7 | 0.79 | 0.69 | 0.64 |
| | 1x100 convolutional kernel | 9 | 0.72 | 0.69 | 0.61 |
| | 4x10 convolutional kernel | 6 | 0.67 | 0.64 | 0.53 |
| | 6-layer operational blocks | 5 | 0.77 | 0.67 | 0.58 |
| | Without filtering | 7 | 0.67 | 0.63 | 0.51 |
| | TR-based data normalization | 10 | 0.51 | 0.51 | 0.45 |
| | Without data normalization | 12 | 0.66 | 0.58 | 0.59 |
| DREAMS | Base model | 7 | 0.71 | 0.62 | 0.62 |
| | Non-weighted cross-entropy | 6 | 0.78 | 0.63 | 0.59 |
| | 1x100 convolutional kernel | 14 | 0.79 | 0.66 | 0.68 |
| | 4x10 convolutional kernel | 10 | 0.72 | 0.61 | 0.61 |
| | 6-layer operational blocks | 8 | 0.81 | 0.63 | 0.62 |
| | Without filtering | 7 | 0.66 | 0.55 | 0.58 |
| | TR-based data normalization | 14 | 0.68 | 0.55 | 0.60 |
| | Without data normalization | 7 | 0.64 | 0.46 | 0.55 |
| ISRUC | Base model | 15 | 0.74 | 0.66 | 0.68 |
| | Non-weighted cross-entropy | 10 | 0.79 | 0.64 | 0.66 |
| | 1x100 convolutional kernel | 17 | 0.76 | 0.68 | 0.69 |
| | 4x10 convolutional kernel | 11 | 0.76 | 0.66 | 0.68 |
| | 6-layer operational blocks | 7 | 0.78 | 0.65 | 0.66 |
| | Without filtering | 17 | 0.76 | 0.68 | 0.69 |
| | TR-based data normalization | 9 | 0.55 | 0.55 | 0.52 |
| | Without data normalization | 11 | 0.56 | 0.54 | 0.54 |

Next, Table 3 shows the results of the second experiment in which each individual (local) model has been used to predict the reference scorings on each of the complete datasets. These individual models resulted from the learning process whose results have been shown in Table 2. Validation results shown in Table 3 therefore involve external dataset evaluations, with the only exception of data in the main diagonal. The main diagonal in Table 3 represents the simulation in which each model is used to predict its own local complete dataset (i.e. not only the corresponding testing set as in Table 2). Consequently, notice that in this case the predictions are biased, as most of the data were already used as part of the corresponding *TR* and *VAL* partitions during the learning process. Nevertheless, it has been decided to include these results in Table 3 for reference and to allow comparison to the corresponding results shown in Table 2.

*Table 3. Performance results for each individual model on predicting the complete datasets when presented as external data. The first column indicates the target dataset to be predicted. For each target dataset the different rows correspond to each of the models' tested architectural variants. The notation M(X) is used to indicate that the model M was trained on the local dataset X. The main diagonal (grey background) shows the results when the model is predicting its own complete local dataset (biased prediction). Results are reported as kappa index with respect to the corresponding reference clinical scorings*

| Target dataset | Model configuration | Individual local models | | | | | |
|---|---|---|---|---|---|---|---|
| | | M(HMC) | M(Dublin) | M(SHHS) | M(Telemetry) | M(DREAMS) | M(ISRUC) |
| HMC | Base model | 0.68 | 0.33 | 0.48 | 0.51 | 0.35 | 0.59 |
| | Non-weighted cross-entropy | 0.72 | 0.34 | 0.56 | 0.46 | 0.23 | 0.52 |
| | 1x100 convolutional kernel | 0.70 | 0.30 | 0.49 | 0.49 | 0.43 | 0.59 |
| | 4x10 convolutional kernel | 0.58 | 0.25 | 0.27 | 0.34 | 0.10 | 0.36 |
| | 6-layer operational blocks | 0.71 | 0.36 | 0.52 | 0.49 | 0.47 | 0.56 |
| | Without filtering | 0.68 | 0.30 | 0.44 | 0.46 | 0.26 | 0.46 |
| | TR-based data normalization | 0.68 | 0.00 | 0.55 | 0.37 | 0.27 | 0.36 |
| | Without data normalization | 0.67 | 0.05 | 0.53 | 0.34 | 0.14 | 0.50 |
| Dublin | Base model | 0.15 | 0.64 | 0.18 | 0.18 | 0.01 | 0.22 |
| | Non-weighted cross-entropy | 0.20 | 0.68 | 0.27 | 0.07 | 0.06 | 0.05 |
| | 1x100 convolutional kernel | 0.09 | 0.65 | 0.17 | 0.03 | 0.03 | 0.04 |
| | 4x10 convolutional kernel | 0.40 | 0.35 | 0.37 | 0.30 | 0.17 | 0.17 |
| | 6-layer operational blocks | 0.08 | 0.64 | 0.09 | 0.07 | 0.03 | 0.12 |
| | Without filtering | 0.53 | 0.65 | 0.24 | 0.44 | 0.20 | 0.38 |
| | TR-based data normalization | 0.00 | 0.58 | 0.00 | 0.00 | 0.00 | 0.01 |
| | Without data normalization | 0.00 | 0.61 | 0.00 | -0.06 | 0.00 | 0.00 |
| SHHS | Base model | 0.62 | 0.34 | 0.68 | 0.56 | 0.49 | 0.66 |
| | Non-weighted cross-entropy | 0.64 | 0.37 | 0.78 | 0.40 | 0.53 | 0.61 |
| | 1x100 convolutional kernel | 0.64 | 0.26 | 0.69 | 0.43 | 0.47 | 0.58 |
| | 4x10 convolutional kernel | 0.60 | 0.03 | 0.68 | 0.44 | 0.19 | 0.37 |
| | 6-layer operational blocks | 0.57 | 0.39 | 0.68 | 0.47 | 0.55 | 0.57 |
| | Without filtering | 0.57 | 0.10 | 0.70 | 0.50 | 0.46 | 0.57 |
| | TR-based data normalization | 0.60 | 0.00 | 0.73 | 0.37 | 0.31 | 0.46 |
| | Without data normalization | 0.63 | 0.04 | 0.72 | 0.37 | 0.17 | 0.48 |
| Telemetry | Base model | 0.57 | 0.18 | 0.42 | 0.69 | 0.28 | 0.60 |
| | Non-weighted cross-entropy | 0.60 | 0.32 | 0.50 | 0.75 | 0.15 | 0.48 |
| | 1x100 convolutional kernel | 0.63 | 0.21 | 0.31 | 0.69 | 0.41 | 0.59 |
| | 4x10 convolutional kernel | 0.45 | 0.17 | 0.12 | 0.46 | 0.09 | 0.09 |
| | 6-layer operational blocks | 0.55 | 0.25 | 0.41 | 0.72 | 0.43 | 0.54 |
| | Without filtering | 0.57 | 0.19 | 0.30 | 0.63 | 0.14 | 0.36 |
| | TR-based data normalization | 0.56 | 0.06 | 0.43 | 0.50 | 0.30 | 0.34 |
| | Without data normalization | 0.52 | 0.09 | 0.39 | 0.63 | 0.20 | 0.46 |
| DREAMS | Base model | 0.33 | 0.29 | 0.36 | 0.40 | 0.67 | 0.65 |
| | Non-weighted cross-entropy | 0.41 | 0.37 | 0.51 | 0.27 | 0.72 | 0.59 |
| | 1x100 convolutional kernel | 0.42 | 0.27 | 0.50 | 0.45 | 0.75 | 0.63 |
| | 4x10 convolutional kernel | 0.44 | 0.20 | 0.33 | 0.17 | 0.64 | 0.35 |
| | 6-layer operational blocks | 0.32 | 0.26 | 0.35 | 0.46 | 0.75 | 0.66 |
| | Without filtering | 0.29 | 0.32 | 0.37 | 0.40 | 0.63 | 0.55 |
| | TR-based data normalization | 0.41 | -0.00 | 0.44 | 0.43 | 0.64 | 0.18 |
| | Without data normalization | 0.49 | 0.03 | 0.43 | 0.46 | 0.59 | 0.41 |
| ISRUC | Base model | 0.63 | 0.34 | 0.61 | 0.56 | 0.54 | 0.72 |
| | Non-weighted cross-entropy | 0.62 | 0.42 | 0.61 | 0.38 | 0.52 | 0.74 |
| | 1x100 convolutional kernel | 0.67 | 0.27 | 0.65 | 0.43 | 0.55 | 0.74 |
| | 4x10 convolutional kernel | 0.63 | 0.41 | 0.57 | 0.45 | 0.21 | 0.69 |
| | 6-layer operational blocks | 0.59 | 0.39 | 0.57 | 0.48 | 0.60 | 0.73 |
| | Without filtering | 0.62 | 0.12 | 0.62 | 0.46 | 0.52 | 0.73 |
| | TR-based data normalization | 0.54 | 0.01 | 0.47 | 0.41 | 0.33 | 0.55 |
| | Without data normalization | 0.54 | 0.08 | 0.49 | 0.35 | 0.25 | 0.55 |

In Table 4, the results of the third experiment (ensemble predictions) are shown and compared to the results achieved by the respective individual models. In Table 4, for each target dataset (first column), the reference local predictions of the different individual models are shown in the third column. These reference levels of local predictions are taken from the performance achieved in their respective local testing sets (last column of Table 2). The fourth column shows the corresponding ranges of inter-database individual predictions as derived from data in Table 3. The resulting average performance is shown in the fifth column. These ranges do not include data from the main diagonal in Table 2, i.e. for dataset $k$, performance of $M(k)$ is excluded. Consequently, results of the fourth and fifth columns exclusively reflect the performance of the individual models' predictions when the dataset is presented as external data. As an example, value in the fifth column of Table 4 with regard to the *base model* configuration for HMC, has been calculated as $0.45 = mean\{0.33, 0.48, 0.51, 0.35, 0.59\}$, derived from the corresponding entry for HMC in Table 3. Finally, the prediction of the ensemble model for the corresponding entry is shown in the last column of Table 4. Similarly, and as stated before in Section 5, *ENS(k)* excludes *M(k)* from the ensemble, e.g. for HMC, kappa values shown in the last column of Table 4 correspond to *ENS[M(Dublin), M(SHHS), M(Telemetry), M(DREAMS), M(ISRUC)]*.

*Table 4. Comparison between the individual models' local and external dataset predictions, and the corresponding ensemble performance. The first column indicates the target dataset to be predicted. Results are reported as kappa index with respect to the corresponding reference clinical scorings. Indices of reference local performance for the individual models are taken from "TS" in Table 2. The corresponding indices of external performance (ranges and average) are taken from data in Table 3.*

| Target dataset | Model configuration | Individual models | | | Ensemble |
|---|---|---|---|---|---|
| | | Reference local performance | External performance | | External performance |
| | | | Range | Average | |
| HMC | Base model | 0.62 | 0.33 − 0.59 | 0.45 | 0.60 |
| | Non-weighted cross-entropy | 0.63 | 0.23 − 0.56 | 0.42 | 0.53 |
| | 1x100 convolutional kernel | 0.65 | 0.30 − 0.59 | 0.46 | 0.57 |
| | 4x10 convolutional kernel | 0.63 | 0.10 − 0.36 | 0.26 | 0.39 |
| | 6-layer operational blocks | 0.61 | 0.36 − 0.56 | 0.48 | 0.58 |
| | Without filtering | 0.61 | 0.26 − 0.46 | 0.38 | 0.49 |
| | TR-based data normalization | 0.62 | 0.00 − 0.55 | 0.31 | 0.51 |
| | Without data normalization | 0.61 | 0.05 − 0.53 | 0.31 | 0.51 |
| Dublin | Base model | 0.44 | 0.01 − 0.22 | 0.15 | 0.19 |
| | Non-weighted cross-entropy | 0.41 | 0.05 − 0.27 | 0.13 | 0.10 |
| | 1x100 convolutional kernel | 0.48 | 0.03 − 0.17 | 0.07 | 0.07 |
| | 4x10 convolutional kernel | 0.33 | 0.17 − 0.40 | 0.28 | 0.40 |
| | 6-layer operational blocks | 0.41 | 0.03 − 0.12 | 0.08 | 0.06 |
| | Without filtering | 0.46 | 0.20 − 0.53 | 0.36 | 0.51 |
| | TR-based data normalization | 0.45 | 0.00 − 0.01 | 0.00 | 0.00 |
| | Without data normalization | 0.43 | -0.06 − 0.00 | -0.01 | -0.06 |
| SHHS | Base model | 0.65 | 0.34 − 0.66 | 0.53 | 0.62 |
| | Non-weighted cross-entropy | 0.70 | 0.37 − 0.64 | 0.51 | 0.61 |
| | 1x100 convolutional kernel | 0.64 | 0.26 − 0.64 | 0.47 | 0.59 |
| | 4x10 convolutional kernel | 0.66 | 0.03 − 0.60 | 0.32 | 0.40 |
| | 6-layer operational blocks | 0.62 | 0.39 − 0.57 | 0.51 | 0.61 |
| | Without filtering | 0.64 | 0.10 − 0.57 | 0.44 | 0.60 |
| | TR-based data normalization | 0.69 | 0.00 − 0.60 | 0.35 | 0.56 |
| | Without data normalization | 0.67 | 0.04 − 0.63 | 0.34 | 0.54 |
| Telemetry | Base model | 0.58 | 0.18 − 0.60 | 0.41 | 0.53 |
| | Non-weighted cross-entropy | 0.64 | 0.15 − 0.60 | 0.41 | 0.55 |
| | 1x100 convolutional kernel | 0.61 | 0.21 − 0.63 | 0.43 | 0.53 |
| | 4x10 convolutional kernel | 0.53 | 0.09 − 0.45 | 0.18 | 0.24 |
| | 6-layer operational blocks | 0.58 | 0.25 − 0.55 | 0.44 | 0.52 |
| | Without filtering | 0.51 | 0.14 − 0.57 | 0.31 | 0.38 |
| | TR-based data normalization | 0.45 | 0.06 − 0.56 | 0.34 | 0.51 |
| | Without data normalization | 0.59 | 0.09 − 0.52 | 0.33 | 0.47 |
| DREAMS | Base model | 0.62 | 0.29 − 0.65 | 0.40 | 0.43 |
| | Non-weighted cross-entropy | 0.59 | 0.27 − 0.59 | 0.43 | 0.50 |
| | 1x100 convolutional kernel | 0.68 | 0.27 − 0.63 | 0.45 | 0.54 |
| | 4x10 convolutional kernel | 0.61 | 0.17 − 0.44 | 0.30 | 0.41 |
| | 6-layer operational blocks | 0.62 | 0.26 − 0.66 | 0.41 | 0.44 |
| | Without filtering | 0.58 | 0.29 − 0.55 | 0.39 | 0.48 |
| | TR-based data normalization | 0.60 | 0.00 − 0.44 | 0.29 | 0.49 |
| | Without data normalization | 0.55 | 0.03 − 0.49 | 0.36 | 0.52 |
| ISRUC | Base model | 0.68 | 0.34 − 0.63 | 0.54 | 0.63 |
| | Non-weighted cross-entropy | 0.66 | 0.38 − 0.62 | 0.51 | 0.58 |
| | 1x100 convolutional kernel | 0.69 | 0.27 − 0.67 | 0.52 | 0.65 |
| | 4x10 convolutional kernel | 0.68 | 0.21 − 0.63 | 0.45 | 0.59 |
| | 6-layer operational blocks | 0.66 | 0.39 − 0.60 | 0.53 | 0.61 |
| | Without filtering | 0.69 | 0.12 − 0.62 | 0.47 | 0.65 |
| | TR-based data normalization | 0.52 | 0.01 − 0.54 | 0.35 | 0.49 |
| | Without data normalization | 0.54 | 0.08 − 0.54 | 0.34 | 0.48 |

Finally, Table 5 shows the global performance results for each of the different tested configurations, averaged across all the datasets. Each row in the second, third and fourth columns of Table 5 is calculated by averaging the corresponding values of columns three, five, and six, respectively, in Table 4. For example, the entry corresponding to the averaged local performance of the *base model* in Table 5 (0.6043 in column two) is calculated as the mean of {0.62, 0.44, 0.65, 0.58, 0.62, 0.68}. Columns five, six and seven in Table 5 respectively represent the averaged inter-database differences between the performances of *(i)* the individual local models in their respective local testing datasets and their averaged external dataset predictions, *(ii)* the individual local models in their respective local testing datasets and the predictions of the ensemble model, and *(iii)* the averaged external dataset predictions of the individual models and the corresponding ensemble model predictions.

*Table 5. Averaged inter-dataset prediction performances for each of the models' configuration variants and comparison among the different prediction scenarios: local testing set using individual models (I), external dataset using individual models (II), and external dataset using an ensemble of individual models (III). The highest absolute values on each column are shown in bold. Results are reported as kappa index with respect to the corresponding reference clinical scorings*

| Model configuration | Individual models - local dataset (I) | Individual models - external datasets (II) | Ensemble - external dataset (III) | I vs II differences | I vs III differences | II vs III differences |
|---|---|---|---|---|---|---|
| Base model | 0.6043 | **0.4133** | 0.5000 | -0.1910 | -0.1043 | 0.0867 |
| Non-weighted cross-entropy | 0.6171 | 0.4017 | 0.4783 | -0.2154 | -0.1388 | 0.0767 |
| 1x100 convolutional kernel | **0.6243** | 0.4000 | 0.4917 | -0.2243 | -0.1326 | 0.0917 |
| 4x10 convolutional kernel | 0.5843 | 0.2983 | 0.4050 | -0.2860 | **-0.1793** | 0.1067 |
| 6-layer operational blocks | 0.5857 | 0.4083 | 0.4700 | -0.1774 | -0.1157 | 0.0617 |
| Without filtering | 0.5771 | 0.3917 | **0.5183** | -0.1854 | -0.0588 | 0.1267 |
| TR-based data normalization | 0.5500 | 0.2733 | 0.4267 | -0.2767 | -0.1233 | **0.1533** |
| Without data normalization | 0.5771 | 0.2783 | 0.4100 | **-0.2988** | -0.1671 | 0.1317 |

## 6.1. Analysis of the experimental results

From a general perspective, the following conclusions might be derived by taking results from Tables 2-5 into consideration:

-Individual local-dataset generalization overestimates the actual inter-dataset generalization for all datasets (I vs II differences in Table 5 are always negative).
-The proposed ensemble method improves the individual inter-dataset generalization performance (II vs III differences in Table 5 show always positive values).
-Individual local generalization still represents an upper bound of the inter-dataset generalization achieved by the ensemble approach (I vs II, and I vs III differences in Table 5 present always negative values).
-Different model architectures and data configuration factors modulate the expected inter-database generalization, regardless of the use of the ensemble approach. These factors are discussed in more detail in the following subsection.

### 6.1.1. Influence of the different tested parameters

The best generalization performance in terms of the individual inter-database predictions (0.4133) was achieved using the base configuration, i.e. the model parameterized with the weighted cross-entropy loss, a 1x10 convolution filter, 3-layer operational blocks, epoch-based input data normalization, and signal filters. This configuration did also achieve the second best prediction using the proposed ensemble approach (0.5000). The best absolute inter-database prediction performance (0.5183) was achieved using the ensemble approach with the configuration variant where no filter pre-processing was applied. In the following we discuss in more detail the influence of the different parameter configurations as observed throughout our experimentation.

#### Weighted vs unweighted cross-entropy:

The use of weighted cross-entropy did not improve the results of the individual models on their corresponding (local) TS datasets (Table 5, average local performance is 0.6043 with weighted cross-entropy, and 0.6171 with normal cross-entropy). When looking at inter-dataset generalization, however, the weighted cross-entropy approach shows better averaged performance for the individual (0.4133 vs 0.4017) and ensemble (0.5000 vs 0.4783) model predictions. In consequence, the relative drop in generalization, from the expected local-database prediction, to the actual individual inter-database prediction, becomes higher when using weighted cross-entropy configuration (Table 5, I vs II differences: -0.2154 vs -0.1910). As per dataset (see Table 4, columns 5 and 6), results were better on HMC, Dublin, SHHS, and ISRUC, while showing lower performance only on DREAMS, and comparable performance on Telemetry. Therefore, our results show that the use of weighted cross-entropy provides better inter-database generalization in general. The result was expected as the particular class distribution in one dataset might not be representative of the corresponding distributions in other datasets.

#### Convolutional kernel size

The use of 2D convolution translates in worse individual inter-database generalization in comparison to the use of 1D convolution. In particular, 2D convolution shows one of the lowest individual averaged generalization performances (0.2983, see I vs II differences in Table 5) with the worst corresponding ensemble generalization (0.4050, see Table 5). When increasing the length of the 1D convolution in the horizontal dimension from 10 (0.1 s) to 100 (1 s) the best local-dataset prediction generalization was achieved (0.6243, column 2 in Table 5), however the result does not hold when looking at the corresponding inter-database generalization values. Hence we interpret this effect as a sign of local dataset overfitting, interestingly, even though the respective local (TS) generalization achieved the best score. According to this result, it seems more effective to establish few restrictions (or *a priori* conditions) on the spatial vicinity of the input feature space, and rather leave the successive network layers to combine the input features guided by the learning process.

Depth of the network

The use of a deeper architecture shows an increase in the training performance (*base model* vs *6-layer depth* configurations comparison in Table 2), but does not show better results regarding any type of generalization (local or inter-dataset) as with respect to the default 3-level architecture (see Table 5). Given this result, we interpret that the model shows enough learning capacity using the 3-level configuration, with no additional benefit by adding extra complexity when increasing the number of extra layers (but rather causing an overfitting effect). Instead, increasing the number of training data might be a better option to boost generalization capabilities of the resulting models.

Data normalization criterion

Our data clearly show that both "Without data normalization" and "TR-based data normalization" configurations achieve worse results in general, as in comparison to the epoch-based normalization procedure. Moreover, the results are the worst achieved among all the tested architectures, both for the individual local- and inter-database results, as well as one of the worst (only exceed by the model using 2D convolution) for the ensemble-based predictions (check columns two, three, and four, respectively, in Table 5). These results highlight the importance of the data normalization strategy as one of the most influential factors to achieve good inter-database generalization. Epoch-based normalization ensures "local" normalization which is database-independent in the sense that the specific normalization does not need to be learned from the training data, and therefore generalizes better across datasets. Notice that the effect is noticeable even if we have incorporated batch-normalization into our network's layer architecture.

Filtering and artifact rejection

Our results do not show a clear trend here. On average, the use of unfiltered signals shows worse local generalization performance among the individual models than when filtered signals are used at the input (see column 2 in Table 5). The main effect here is due to the contribution of the results achieved in the Telemetry and DREAMS datasets (see Table 2). Unfiltered signals resulted into worse generalization capabilities as well, when taking as reference the global results among the individual model's predictions (I vs II differences in Table 5). Remarkably, however, the best ensemble-based generalization score was achieved when signal filtering was not applied (0.5183, column 4 in Table 5).

When looking at the inter-database predictability of each dataset (see Table 3, and columns 4 and 5 in Table 4) we see that the use of filters improves predictability in general, but with the notable exception of the Dublin dataset. From Table 3 it is also evident that the model based on the Dublin data has the most difficulties, in general, to predict the rest of the datasets. Apparently, the filtering effect seems to have a totally different trend on the Dublin dataset as compared to the rest. The poor generalized performance when predicting the filtered Dublin dataset likely negatively influences the ensemble too, given ensemble's performance rises significantly when unfiltered data are used. Separated tests have been conducted (omitted from the results for simplicity) showing that by removing Dublin from the ensemble, performance increases, which supports the filtering option. In other words, we do not attribute the best ensemble result to the convenience of using unfiltered data itself, but to the unique characteristics of the Dublin dataset, and the outlier effect of including M(Dublin) in the ensemble. Specific discussion on the results achieved in the Dublin dataset can be found in the Supplementary material.

In conclusion, although the use of filtered data seems to have an overall positive effect in the resulting model's generalization, this effect does not seem to be reproducible across all datasets, and therefore we rather brand the results regarding the use of signal filtering as inconclusive.

Comparative analysis with other studies in the literature

In Table 6 validation results comprising other automatic approaches referenced in this work are summarized. Results are structured considering if the performance metrics were obtained on the basis of a local or an external dataset validation scenario, as discussed throughout this work. Only validation metrics that report agreement in terms of Cohen's kappa are considered for comparison purposes.

*Table 6. Summary of validation results from other reference works on automatic sleep staging. When available, performance and agreement metrics are reported in terms of Cohen's kappa index. When validation does not report Cohen's kappa the term "(no kappa)" is used. If no validation data at all is reported then N/A (Not Available) is indicated. Results regarding our study take as reference the "base model" architecture configuration as described in the previous tables. For each database, the number of recordings composing the dataset is indicated between brackets, as well as its accessibility (internal/private, or public/upon request).*

| Study | Database | Deep Learning | Local prediction scenario | External prediction scenario | Expected human agreement |
|---|---|---|---|---|---|
| Kemp et al. [4] | In-house (6, internal) | No | (no kappa) | N/A | (no kappa) |
| Principe et al. [5] | In-house (5, internal) | No | (no kappa) | N/A | N/A |
| Agarwal et al. [6] | Sleep laboratory at the Hôpital Sacré Coeur (HSC), Montreal, Canada (12, internal) | No | (no kappa) | N/A | (no kappa) |
| Estevez et al. [7] | Sleep Laboratory of the Instituto de Nutrición y Tecnología de Alimentos, INTA, Universidad de Chile (11, internal) | No | 0.84 | N/A | N/A |
| Anderer et al. [8] | SIESTA (72, public**) | No | 0.75 | N/A | 0.76 (see [27]) |
| Jo et al. [9] | In-house (4, internal) | No | (no kappa) | N/A | N/A |
| Pittman et al. [10] | Clinical sleep laboratory of Brigham and Women's Hospital (31, internal) | No | 0.61 – 0.67 | N/A | 0.73 |
| Alvarez-Estevez et al. [11] | SHHS (33, public**) | No | 0.77 | N/A | 0.81-0.83 (see [30]) |
| Khalighi et al. [12] | ISRUC (40, public) | No | (no kappa) | N/A | 0.87 (see [49]) |
| Stepnowsky et al. [19] | New York University (NYU) School of Medicine (23, internal) | No | N/A | 0.59 - 0.63 | 0.77-0.89 |
| | New York University (NYU) School of Medicine and Sleep Medicine Associates of New York City (21, internal) | No | N/A | 0.42 - 0.63 | 0.46-0.80 |
| Hassan et al. [13] | Sleep-EDF (8, public) | No | 0.86 | N/A | N/A |
| | DREAMS (20, public) | No | (no kappa) | N/A | N/A |
| Tsinalis et al. [14] | Sleep-EDF (20, public) | Yes | (no kappa) | N/A | N/A |
| Dong et al. [15] | Montreal Archive of Sleep Studies (MASS) (62, public**) | Yes | (no kappa) | N/A | N/A |
| Supratak et al. [16] | Sleep-EDF (20, public) | Yes | 0.76 | N/A | N/A |
| | Montreal Archive of Sleep Studies (MASS) (62, public**) | Yes | 0.80 | N/A | N/A |
| Zhang et al. [17] | Dublin (25, public) | Yes | 0.84 | N/A | N/A |
| Fernández-Varela et al. [18] | SHHS (1000, public**) | Yes | 0.83 | N/A | 0.81-0.83 (see [30]) |
| Bresch et al. [20] | In-house (147, internal) | Yes | 0.73 | 0.70 | N/A |
| | SIESTA (588, public**) | Yes | 0.76 | 0.45 | 0.76 (see [27]) |
| Stephansen et al. [21] | IS-RC (70, internal) | Yes | N/A | 0.72 – 0.77 | 0.58 |
| Malafeev et al. [22] | Sleep laboratory of the Institute of Pharmacology and Toxicology at the University of Zurich (54, internal) | Yes | 0.83* | N/A | N/A |
| | Sleep Disorders Center, Department of Clinical Neurophysiology, Institute of Psychiatry and Neurology in Warsaw, Warsaw, Poland (22, internal) | Yes | N/A | 0.41-0.61* | N/A |
| Our study | HMC (159, internal) | Yes | 0.62 | 0.60 | N/A |
| | Dublin (25, public) | Yes | 0.44 | 0.19 | N/A |
| | SHHS (100, public**) | Yes | 0.65 | 0.62 | 0.81-0.83 [30] |
| | Telemetry (44, public) | Yes | 0.58 | 0.53 | N/A |
| | DREAMS (20, public) | Yes | 0.62 | 0.43 | N/A |
| | ISRUC (100, public) | Yes | 0.68 | 0.63 | 0.87 [49] |

*\* Inter-stage (W, N1-N3, R) kappa average of best performing method in the local testing scenario (1p2_32u_128ep and 1p2_32u_32ep, see Supplementary Materials in [22] for details)*
*\*\*Access granted upon request or subject to evaluation procedure*

## 7. Discussion

This work addresses some of the problems associated with the development of ML models with robust generalization capabilities on common-task multiple-database scenarios. Focusing on a specific medical domain (sleep staging in sleep medicine) we have shown the non-triviality of translating the estimated model's local generalization capabilities into independent external databases. Our results in this regard are conclusive, showing a consistent overestimation trend in the performance achieved by the local models when compared with their corresponding external dataset predictions. A positive aspect of our work is that the datasets used are almost entirely public (with the only exception of HMC). This fact makes replication and comparison of the results possible in future works.

Validation procedures regarding automatic sleep staging reported so far in the literature are limited. Usually validations are performed using only one dataset, or lacking the correct database-wise independent separation of the data, needed to assess the real generalization capabilities across multiple databases. Hence, our findings suggest the related scientific literature should be critically reviewed. Surprisingly enough, new works continue to appear in scientific journals claiming "good generalization performance". Nevertheless, these studies are missing the proper experimental design to support such a statement. To our knowledge, only a few works [19] [20] [21] [22] have included performance generalization tests using independent external databases as part of their experimentation. In Malafeev et al. [22] a clear downgrade in performance is noticeable when comparing the results from their local database validation (Tables S1 and S2, test dataset in Malafeev et al. [22]) with the corresponding generalization results using an external dataset (Tables S3 and S4 in Malafeev et al. [22]), thus supporting our findings. A similar pattern is found in the experiments conducted in Bresch et al. [20], which showed a significant performance drop from $\kappa = 0.70$ to $\kappa = 0.45$ when swapping the two independent training and testing databases. Results are more difficult to interpret in Stephansen et al. [21] and Stepnowsky et al. [19]. In the first study, while $\kappa = 0.72 - 0.77$ was reported for the best model on an independent external dataset (IS-RC, see Table 1 in Stephansen et al. [21]), a comparable evaluation of the same model using a local validation setting was not included among the reported results. It is not possible, therefore, to evaluate possible differences between the respective local and external database generalizations. Data in Table 2 in the same work, on the other hand, point out to significant variability effects among different local testing datasets, hence, in line with the general trend reported here and in other works. Similar to Stephansen et al. [21], in Stepnowsky et al. [19] validation performance of $\kappa = 0.42 - 0.63$ is reported on an external dataset, however, the reference local generalization performance of the system is missing from the published results.

Proof of database-independent generalization capabilities is a critical aspect on the eventual acceptability and usability of ML models in real (clinical) practice. Little interest or practical utility (if any) could be attributed to a system that only provides good results on its own local database. One can imagine a patient whose diagnostic test will only be valid within the hospital where the test is executed, with no possibility to achieve comparable conclusions when the same patient is examined by a specialist from another center.

That said, the previous statement should be carefully examined. It is obviously not an unusual practice within the medical field to submit a diagnosis for a second opinion (with the possibility of reaching a different conclusion). Recall from Table 1, one of the factors hampering the inter-database generalization capabilities of ML models in this context (but in general, in any context for which the standard reference is subject to human interpretation) is the inherent subjectivity associated with human decision-making. While this factor equally affects both human and computer-based decision scenarios, the general assumption is that human experts possess enough abstraction capacity that the chances of disagreement are usually low (i.e. they represent the exception and not the rule). In other words, and following with the clinical example, the actual level of disagreement is acceptable for patients to continue trusting the medical system. The key to the matter, therefore, is to set the algorithm's inter-database performance goal to the same level of agreement that would be expected from different human-experts in the context of the same task.

Specifically, for the sleep scoring task, literature has reported human-expert agreement indices between $\kappa = 0.42 - 0.89$, depending on the study [10] [30] [19] [55] [26] [27] [28] [31] [49]. Results are not conclusive, but suggest that agreement for scorers from the same center ($\kappa = 0.73 - 0.87$, [10] [30] [49]) present less variability than when scorers belong to different centers ($\kappa = 0.46$-$0.89$, [19] [55] [26] [27] [28]). Interestingly, the wide range of agreements suggests a strong study-dependency component. Consequently, evaluation of a model's performance acceptability should necessarily be linked to the specific human expected agreement, which is local to the examined dataset. Indeed, it is conceivable that human experts find certain datasets more challenging than others. For example, if the dataset contains a relatively high presence of artifacts, one should also expect a higher level of human variability. On the other hand, the low human-expert agreement achieved in some of the studies might also reveal the necessity to review the current medical scoring standards. It is plausible that potential disagreements are due to a lack of clarity regarding the reference scoring rules, and thus, clearer definitions might improve the overall repeatability of human sleep staging, in particular for the scoring of stages N1 and N3 [31].

Unfortunately, not many works in the automatic sleep staging literature have reported the expected levels of human agreement associated with their testing datasets. A possible reason for this is related to the general low availability of clinical experts and the high costs associated with the development of such experiments. Some exceptions are the works of Pittman et al. [10] and Stepnowsky et al. [19], which achieved slightly lower agreement ($\kappa = 0.61$-$0.67$ [10], $\kappa = 0.42$-$0.63$ [19]) than their respective human references ($\kappa = 0.73$ [10], $\kappa = 0.46$-$0.89$ [19]). More examples can be found for which the reference human agreement can be derived from independent studies using the same database, but that target a different subset of recordings [8] [11] [12] [18] [20]. This is also the for SHHS in this study. Notice that all the previous cases belong to public databases. Exceptionally, Stephansen et al. [21] and Stepnowsky et al. [19] are, to our knowledge, the only works reporting results on an external independent dataset, both for the human scorers and for the scoring algorithm. Arguably, experimental data in Stephansen et al. [21] suggests that the scoring algorithm apparently outperforms the individual human scorers when comparing the results versus a human-expert consensus (see Table 1 in Stephansen et al. [21]).

Regarding the datasets used in this study, human agreement has been reported for the ISRUC ($\kappa = 0.87$ [49]) and for the SHHS ($\kappa = 0.81$-$0.83$ [30]) databases. As stated before, we note that in the case of SHHS, the specific subset and number of recordings used in Whitney et al. [30] ($n = 30$) differs from those used here ($n = 100$). In either case, comparison to the respective performance achieved by the models developed in this work continues to suggest there is room for improvement. In particular, for ISRUC, the best individual model achieves a local predictability of $\kappa = 0.69$, whereas the external prediction resulted in $\kappa = 0.65$ (see columns 3 and 6 of Table 4, respectively). For SHHS, on the other hand, the corresponding local and external generalization performances were of $\kappa = 0.70$ and $\kappa = 0.62$. These results are not surprising, as in this work the focus was not set on the development of an optimal solution for the automatic sleep staging problem. Rather, the main goal was to deal with the database generalization problem using sleep staging as an illustrative applied domain. In this regard, possible improvements are later discussed as future work. To our knowledge, no studies have been performed that analyze the expected human variability for the rest of the datasets included in this work.

Once the database variability problem has been verified, and having discussed its consequences, the next important task is to decide how to deal with it. Throughout this work, we have analyzed some of the scalability problems associated with the methodology that combines different source datasets as input to a unique learning model. In an attempt to cope with these limitations, a strategy based on the training of local models has been proposed, which are flexibly combined in the form of an ensemble. One key advantage of the proposed approach is that it allows flexible reconfiguration and scalability, without the need to retrain the previous model for each new incoming dataset. Our simulations have also shown that the ensemble strategy obtains better inter-database generalization results in comparison to the performance achieved by the use of individual local models.

The idea of using ensembles actually represents a straightforward solution to approach the multi-database learning problem. First, it offers a natural parallelism with the traditional way by which expert disagreements are approached in decision theory and reliability studies. For example, when different expert classifications are available for the classification of an object, usually the valid reference is established by developing some sort of "consensus scoring". Not accidentally, when assuming that each expert's criterion is equally valid, a consensus is usually established by taking the majority vote as the prevailing reference [56] [57]. In fact, each individual ML model can be reinterpreted as a "local expert" that mimics the particular general and ad-hoc knowledge of the human experts on the specific source dataset. In addition, there are several statistical, computational, and representational reasons for, supporting the use of classifier ensembles to address our current problem setting [58] [59]. It might be particularly interesting to contrast our approach with the so-called *bagging* strategy [60], which is widely known to be an effective technique in ML. In this regard, we can intuitively identify each of the individual source datasets with the bootstrap replicates generated from the underlying base dataset, as proposed in the bagging method. Effectively, we can abstract this base dataset from the common feature space representing the underlying goal task (in our case, sleep staging). The necessary diversity to make the ensemble work is reasonably guaranteed because of the different sources of variability and uncertainty associated with each of the source datasets (recall, once again, Table 1). Further discussion about ensemble methods and the underlying principles motivating the majority vote strategy are given in work of Kuncheva [61].

Our experimentation has also shown that not only the network architecture, but how data are pre-processed, both have important consequences on the inter-database generalization capabilities of the resulting models. A detailed analysis of the different tested variants has been carried out in the preceding sections. In particular, the data normalization strategy seems to notably affect the resulting model´s generalization capacity.

Possible limitations of our study should be discussed as well. First, as stated before our validation results still point out toward general suboptimal performance of the automatic models developed in this work when compared to the corresponding expected levels of human scoring variability. Again, this is not surprising, given that important domain implications have been omitted in the development of our models. In particular, the effect of the epoch sequence on the scoring was not considered; that is, the fact that the decision on the classification of the current epoch is partially influenced by the sleep state of the preceding and subsequent epochs. Future developments dealing with sequence learning, by adding extra layers in combination with Recurrent Neural Networks (e.g. Long Short-Term Memory or similar) should improve the overall performance. In fact, some of the recent studies already applied this idea [20] [22] [19] [15] [16]. Better model's hyper-parameterization and data pre-processing need to be explored as well. In particular the use of alternative source signal derivations and a different base sampling rate (currently 100 Hz) is of interest. As discussed in Section 2, many sleep microstructure events clearly manifest among specific signal derivations and thus can influence inter-database generalization results. Similarly, resampling the input signals to a common rate of 100 Hz limits the working frequency to 50 Hz, which is rather low to capture all EMG muscle activity of interest. In this respect, however, we are limited by the original montages and sampling rates used in the source databases (as low as 64 Hz in some cases, see Supplementary Table 1 for more details). Further, the effects of the proposed filtering pipeline remain unclear, and variability of the results with regard to the Dublin dataset needs to be studied in more detail. Finally, more research is needed in order to confirm the adequacy of the proposed ensemble approach. Future research will also include the exploration of alternative ensemble combining strategies that might outperform the majority vote. For example, a Naive-Bayes combiner [61] seems to be an appealing approach given the different output probability distributions associated with each of the individual models in the ensemble.

In conclusion, we have shown that validation of the generalization performance of an ML model on an independent local dataset is not adequate proof of its true generalization capabilities. When confronting it with external datasets, even with regard to a shared common task, the actual prediction performance is likely to be downgraded. Scientific literature which claims good generalization capabilities, therefore, should be critically reviewed taking this point into consideration. The use of an ensemble of local models appears to be an interesting approach because of its advantages regarding data escalation and the overall improved generalization of the resulting model. More studies are needed, however, to confirm this hypothesis. We have also shown the generalization capabilities of a model are not only associated with the architecture of the model itself, or to its parameterization, but also with the careful preprocessing of the input data. In particular, the data normalization strategies, and their granularity, seem to notably affect the generalization capacities of the resulting models.

### Acknowledgments
This research did not receive any specific grant from funding agencies in the public, commercial, or not-for-profit sectors

## Supplementary Material

### A. Combining data from different source datasets as input for machine learning

Here we explore the complexity associated with the exploration of the different input combinations when data from different datasets are available for training of a machine learning model. A reason for this exploration might be to find the combination leading to best model in terms of generalization capabilities. For this purpose we are assuming a classical data partitioning schema [1] in which data from each dataset $X$ are split into $X_{TR}$, $X_{VAL}$, and $X_{TS}$ partitions, respectively, for the purposes of training ($TR$), validation ($VAL$), and testing ($TS$) of a model. Notice that, in the case of neural networks, the validation subset can be used as well to implement the early stopping criterion [1]. Let us also assume that, in total, $N$ independent datasets are available as input for training purposes, and that all the corresponding $X_{TS}$ subsets are set aside as independent testing data. Without loss of generality, we are leaving out of the discussion the specific proportion of $X$ data assigned to each of the $X_{TR}$, $X_{VAL}$, and $X_{TS}$ subsets.

Under these circumstances, the task of exploring the possible input combinations translates into how data from the different $X_{TR}$, $X_{VAL}$ subsets ($2N$ in total) can be distributed into individual ($TR$, $VAL$) pairs. To illustrate that, let us consider the simple case in which $N=2$, and list all the resulting combinations in Supplementary Table A1. Each dataset is identified sequentially in alphabetical order (i.e. for $N = 2$, datasets are denoted $A$ and $B$).

*Supplementary Table A1. Possible input combinations when N=2*

| n | TR | VAL | n | TR | VAL |
|---|---|---|---|---|---|
| 1 | $A_{TR}$ | - | 7 | $B_{TR}$ | $B_{VAL}$ |
| 2 | $A_{TR}$ | $A_{VAL}$ | 8 | $B_{TR}$ | $A_{VAL},B_{VAL}$ |
| 3 | $A_{TR}$ | $B_{VAL}$ | 9 | $A_{TR},B_{TR}$ | - |
| 4 | $A_{TR}$ | $A_{VAL},B_{VAL}$ | 10 | $A_{TR},B_{TR}$ | $A_{VAL}$ |
| 5 | $B_{TR}$ | - | 11 | $A_{TR},B_{TR}$ | $B_{VAL}$ |
| 6 | $B_{TR}$ | $A_{VAL}$ | 12 | $A_{TR},B_{TR}$ | $A_{VAL},B_{VAL}$ |

According to Supplementary Table A1, a total of 12 combinations are possible for $N=2$. The case ($X_{TR}$, -) represents the situation where no validation set was used during training of the model. Notice that, for simplicity, we are assuming that the $X_{TR}$ and $X_{VAL}$ subsets are fixed, and that only $X_{TR}$ subsets can be assigned to $TR$, and likewise, that only $X_{VAL}$ subsets can be assigned to $VAL$. We are thus omitting possibilities such as ($X_{TR}$ + $X_{VAL}$, -) or ($X_{VAL}$, $X_{TR}$). Obviously, by considering such possibilities, the associated combinatorial input space would grow even more.

We can now do a similar exercise for the extended case $N = 3$. The resulting combinations are subsequently shown in Supplementary Table A2.

*Supplementary Table A2. Possible input combinations when N=3*

| n | TR | VAL | n | TR | VAL | n | TR | VAL |
|---|---|---|---|---|---|---|---|---|
| 1 | $A_{TR}$ | - | 17 | $C_{TR}$ | - | 33 | $A_{TR},C_{TR}$ | - |
| 2 | $A_{TR}$ | $A_{VAL}$ | 18 | $C_{TR}$ | $A_{VAL}$ | 34 | $A_{TR},C_{TR}$ | $A_{VAL}$ |
| 3 | $A_{TR}$ | $B_{VAL}$ | 19 | $C_{TR}$ | $B_{VAL}$ | … | … | … |
| 4 | $A_{TR}$ | $C_{VAL}$ | 20 | $C_{TR}$ | $C_{VAL}$ | 40 | $A_{TR},C_{TR}$ | $A_{VAL},B_{VAL},C_{VAL}$ |
| 5 | $A_{TR}$ | $A_{VAL},B_{VAL}$ | 21 | $C_{TR}$ | $A_{VAL},B_{VAL}$ | 41 | $B_{TR},C_{TR}$ | - |
| 6 | $A_{TR}$ | $A_{VAL},C_{VAL}$ | 22 | $C_{TR}$ | $A_{VAL},C_{VAL}$ | 42 | $B_{TR},C_{TR}$ | $A_{VAL}$ |
| 7 | $A_{TR}$ | $B_{VAL},C_{VAL}$ | 23 | $C_{TR}$ | $B_{VAL},C_{VAL}$ | … | … | … |
| 8 | $A_{TR}$ | $A_{VAL},B_{VAL},C_{VAL}$ | 24 | $C_{TR}$ | $A_{VAL},B_{VAL},C_{VAL}$ | 48 | $B_{TR},C_{TR}$ | $A_{VAL},B_{VAL},C_{VAL}$ |
| 9 | $B_{TR}$ | - | 25 | $A_{TR},B_{TR}$ | - | 49 | $A_{TR},B_{TR},C_{TR}$ | - |
| 10 | $B_{TR}$ | $A_{VAL}$ | 26 | $A_{TR},B_{TR}$ | $A_{VAL}$ | 50 | $A_{TR},B_{TR},C_{TR}$ | $A_{VAL}$ |
| 11 | $B_{TR}$ | $B_{VAL}$ | 27 | $A_{TR},B_{TR}$ | $B_{VAL}$ | 51 | $A_{TR},B_{TR},C_{TR}$ | $B_{VAL}$ |
| 12 | $B_{TR}$ | $C_{VAL}$ | 28 | $A_{TR},B_{TR}$ | $C_{VAL}$ | 52 | $A_{TR},B_{TR},C_{TR}$ | $C_{VAL}$ |
| 13 | $B_{TR}$ | $A_{VAL},B_{VAL}$ | 29 | $A_{TR},B_{TR}$ | $A_{VAL},B_{VAL}$ | 53 | $A_{TR},B_{TR},C_{TR}$ | $A_{VAL},B_{VAL}$ |
| 14 | $B_{TR}$ | $A_{VAL},C_{VAL}$ | 30 | $A_{TR},B_{TR}$ | $A_{VAL},C_{VAL}$ | 54 | $A_{TR},B_{TR},C_{TR}$ | $A_{VAL},C_{VAL}$ |
| 15 | $B_{TR}$ | $B_{VAL},C_{VAL}$ | 31 | $A_{TR},B_{TR}$ | $B_{VAL},C_{VAL}$ | 55 | $A_{TR},B_{TR},C_{TR}$ | $B_{VAL},C_{VAL}$ |
| 16 | $B_{TR}$ | $A_{VAL},B_{VAL},C_{VAL}$ | 32 | $A_{TR},B_{TR}$ | $A_{VAL},B_{VAL},C_{VAL}$ | 56 | $A_{TR},B_{TR},C_{TR}$ | $A_{VAL},B_{VAL},C_{VAL}$ |

Observe that, following the same schema, a total of 56 possible combinations result for *N*=3. In general, given *N* datasets are available, the total number of combinations (#comb) of this manner results:

$$\#comb = (2^N - 1)2^N$$

Intuitively, the formula can be derived by considering that for each element of the ($X_{TR}$, $X_{VAL}$) duple, each of the *N* datasets can be chosen or not, thus $2^N$, but in the first case ($X_{TR}$) the empty combination is not allowed, hence ($2^N - 1$), obviously, as the training dataset cannot be empty.

If considering the case in which data from the *N* datasets are used as input to train one single machine learning model, it is easy to see that this formula represents, not only the complexity of the combinatorial search space, but also the complexity of the training space. Indeed, as each combination involves training of a new model. If we denote these spaces respectively as $\Omega_{SC}$ and $\Omega_{ST}$, where subscript "*S*" references the "single" model approach, it also follows from the formula above that $\Omega_{SC} = \Omega_{ST} = O(2^{2N})$, thus resulting in a class of exponential complexity.

On the other hand, we can consider the scenario in which data combinations occur by means of an ensemble of local models. In that case, under the condition that each individual model *M*(*X*) results only from data in the local dataset *X*, then the (*TR, VAL*) partitioning combinations are limited to either ($X_{TR}$, -) or ($X_{TR}$, $X_{VAL}$). Consequently, under these circumstances it is easy to see that the complexity of the training space results $\Omega_{EC} = 2N$, i.e. linear complexity, where the subscript "*E*" references the ensemble approach.

To deduce the corresponding formula for $\Omega_{ET}$, let us first rewrite the corresponding possible combinations for *N*=3. The resulting list is shown in the Supplementary Table A3.

*Supplementary Table A3. Possible input combinations for N=3 under the ensemble approach*

| n | TR | VAL | n | TR | VAL | n | TR | VAL |
|---|----|-----|---|----|-----|---|----|-----|
| 1 | $A_{TR}$ | - | 10 | $A_{TR}B_{TR}$ | $A_{VAL},B_{VAL}$ | 19 | $A_{TR},B_{TR},C_{TR}$ | - |
| 2 | $A_{TR}$ | $A_{VAL}$ | 11 | $A_{TR},C_{TR}$ | - | 20 | $A_{TR},B_{TR},C_{TR}$ | $A_{VAL}$ |
| 3 | $B_{TR}$ | - | 12 | $A_{TR},C_{TR}$ | $A_{VAL}$ | 21 | $A_{TR},B_{TR},C_{TR}$ | $B_{VAL}$ |
| 4 | $B_{TR}$ | $B_{VAL}$ | 13 | $A_{TR},C_{TR}$ | $C_{VAL}$ | 22 | $A_{TR},B_{TR},C_{TR}$ | $C_{VAL}$ |
| 5 | $C_{TR}$ | - | 14 | $A_{TR},C_{TR}$ | $A_{VAL}, C_{VAL}$ | 23 | $A_{TR},B_{TR},C_{TR}$ | $A_{VAL},B_{VAL}$ |
| 6 | $C_{TR}$ | $C_{VAL}$ | 15 | $B_{TR},C_{TR}$ | - | 24 | $A_{TR},B_{TR},C_{TR}$ | $A_{VAL}, C_{VAL}$ |
| 7 | $A_{TR}B_{TR}$ | - | 16 | $B_{TR},C_{TR}$ | $B_{VAL}$ | 25 | $A_{TR},B_{TR},C_{TR}$ | $B_{VAL},C_{VAL}$ |
| 8 | $A_{TR}B_{TR}$ | $A_{VAL}$ | 17 | $B_{TR},C_{TR}$ | $C_{VAL}$ | 26 | $A_{TR},B_{TR},C_{TR}$ | $A_{VAL},B_{VAL},C_{VAL}$ |
| 9 | $A_{TR}B_{TR}$ | $B_{VAL}$ | 18 | $B_{TR},C_{TR}$ | $B_{VAL}, C_{VAL}$ | --- | --- | --- |

Effectively, it turns out that for *N*=3 the number of combinations reduces to 26 (instead of 56 as in the single model approach). A reason for that is that cases such as ($A_{TR}$, $B_{VAL}$) are not allowed, given the condition that each model should be kept local to its data. Similarly, combinations such as ($A_{TR}B_{TR}$, $A_{VAL}C_{VAL}$) are not possible anymore. Notice, on the other hand, that combinations such as ($A_{TR}B_{TR}$, $A_{VAL}B_{VAL}$) still can be achieved, for example, by combining the individual models resulting from ($A_{TR}$, $A_{VAL}$) and ($B_{TR}$, $B_{VAL}$) into the ensemble. By taking this into account, and comparing Supplementary Table A2 and Supplementary Table A3, it is easy to see that $\Omega_{SC} > \Omega_{EC}$.

In general, for *N* datasets, the corresponding combinational search space following the proposed ensemble approach results:

$$\Omega_{EC} = \sum_{k=1}^{N} \binom{N}{k} \left[ \sum_{j=1}^{k} \binom{k}{j} + 1 \right]$$

While derivation might be less direct, intuitively each *k* term of the outer summation makes reference to the corresponding possible combinations of *k* elements for the *TR* part, given the imposed restrictions. For each of these combinations, the internal summation in *j*, similarly references the possible combinations for the resulting *VAL* elements, including the possibility in which the validation subset is empty (therefore + 1).

On the other hand, while $\Omega_{SC} > \Omega_{EC}$, it is possible to show that $\Omega_{EC}$ still represents exponential complexity. Indeed, by taking the last term of the outer summation into consideration, and operating it a bit:

$$\binom{N}{N} \left[ \sum_{j=1}^{N} \binom{N}{j} + 1 \right] = \sum_{j=1}^{N} \binom{N}{j} + 1 = \sum_{j=0}^{N} \binom{N}{j}$$

one can apply the following result derived from the Binomial Theorem:

$$\sum_{j=0}^{N} \binom{N}{j} = \binom{N}{0} + \binom{N}{1} + \binom{N}{2} + \cdots + \binom{N}{N} = 2^N$$

Hence, in effect $\Omega_{EC} = O(2^N)$.

Thus, as a final remark, notice that while both the combinational and the training spaces have exponential complexity for the case in which one single model is trained, when using the proposed ensemble approach, even though the combinational space still remains exponential, the effective training space significantly reduces from exponential to linear complexity.

## B. Discussion about the results in the Dublin dataset

This dataset is peculiar for several reasons. Indeed, experimental data show that this is usually the most difficult dataset to be predicted by the external models (see Table 3 and Table 4) but also by the model based on Dublin data itself (while *TR* performance is comparable to other models in Table 2, the corresponding *TS* performance is notably decreased). However, in contrast with the general trend among the rest of the datasets, predictability of Dublin by other models apparently improves when no signal filters are applied (in Table 4, individual generalization of 0.15 vs 0.36, and ensemble generalization of 0.19 vs 0.51, respectively, with and without the use of filters). At the same time, however, a similar effect cannot be observed when considering the local generalization capabilities of the model based on Dublin data (see Table 2, 0.44 vs 0.46 in TS). Simulations in which the different filtering steps were individually excluded from the filtering pipeline (not included in the results for simplicity) showed overall downgraded predictability, hence for all the datasets including Dublin. For some reason, however, the overall combination of filters seems to negatively affect Dublin's data predictability, not contributing to the expected homogenization effect as for other databases. Notably, the different models derived from the Dublin data have shown the most difficulties to predict the rest of the datasets in general (see Table 3). While we are not able to find a convincing explanation for these effects, several of the following factors might be contributing to the observed results:

-First, Dublin is the only dataset for which the signals have been digitalized using normalized units, i.e. physical dimensions are not specified in μV, but on normalized volts (NV) (see Supplementary Table 1). However, no information on how this normalization was performed is available on the reference documentation [2]. Physical values in the EDF header are set to minimum and maximum values of 0 and 1 respectively, but after scaling up the digital values (see EDF specifications for details [3]), it turns out that physical (normalized) values go beyond this interval (to around ±3 NV). When compared to the rest of the datasets, the physical values show ranges of ±30 μV at least. Recall from our results that the normalization method has considerable influence on the inter-database predictability of the datasets. In particular, experimental data have shown an overall positive effect of the epoch-based normalization method over simulations where input data normalization was not applied. Consequently, differences due to the normalization factors might indeed explain, at least in part, the results achieved with Dublin.

-Second, Dublin is the dataset with the lowest sampling rate for the EMG, abnormally low (64 Hz) which leads to a workable range of 0-32 Hz. Taking into account that most of the interesting information for the EMG starts over 15-20 Hz, such a sampling rate translates into seldom 20-32 Hz workable range in practice. Furthermore, notice that due to the small sampling rate, the Notch filtering (50 or 60 Hz) is not having any influence on removing possible mains interference from the EMG. A possible artificial effect due to the oversampling of the signal up to 100 Hz is in principle discarded, as it should equally affect the filtered and the unfiltered version of the data, which is not the case. We would also discard an effect due to differences in the DC offset of the EMG signal, as the HP filter should effectively contribute to minimize this factor.

*Supplementary Table 1. Summary characteristics of the datasets included in the experimentation*

| Dataset | Source | Acquisition device | Used derivations | Sampling Rate (in Hz) | Pre-filtering (in Hz) | Mains frequency (in Hz) | Physical range |
|---|---|---|---|---|---|---|---|
| HMC | Sleep Center, HMC Haaglanden, The Netherlands | SOMNOscreen Plus and 10-20 (SOMNOmedics, Germany) | EEG1 O2-Cz, C4-M1<br>EEG2 Fpz-Cz<br>EOG E2-E1<br>EMG submental (bipolar)<br><br>ECG modified lead-II | 256<br>256<br>256<br>256<br><br>256 | 0.2 – 128, 0.2 - 35<br>0.2 – 128, 0.2 - 35<br>0.2 – 128, 0.2 - 35<br>0.2 – 1000, 0.2 - 150<br><br>0.2 - 1000, 0.2 - 150 | 50 | ±800 µV<br>±800 µV<br>±800 µV<br>±800 µV<br><br>±2400 µV |
| Dublin | Sleep Disorders Clinic at St Vincent's University Hospital, Dublin | Jaeger-Toennies system (Erich Jaeger GmbH, Germany) | EEG1 C3-A2<br>EEG2 C4-A1<br>EOG Left-Right<br>EMG submental<br><br>ECG modified lead-II | 128<br>128<br>64<br>64<br><br>128 | 0.30 – 35<br>0.30 – 35<br>0.30 – 35<br>10 – 75<br><br>0.30 - 75 | 50 | — NV (Normalized Volts) |
| SHHS | Sleep Health Heart Study (SHHS) multi-center cohort, USA | Compumedics P-series Sleep Monitoring system, versions 3-4, (Compumedics Limited, Australia) | EEG1 C3-A2<br>EEG2 C4-A1<br>EOG Left-Right<br>EMG submental (bipolar)<br><br>ECG modified lead-II | 125/128<br>125/128<br>50/64<br>125/128<br><br>250/256 | 0.15 – N/A<br>0.15 – N/A<br>0.15 – N/A<br>0.15 – N/A<br><br>0.15 – N/A | 60 | ±125 µV<br>±125 µV<br>±125 µV<br>±31.5 µV<br><br>±1.25 mV |
| Telemetry | Leiden University Hospital, The Netherlands | Telemetry system (Kemp et al. [1], The Netherlands) | EEG1 Pz-Oz<br>EEG2 Fpz-Cz<br>EOG horizontal<br>EMG submental<br><br>N/A (no ECG recorded) | 100<br>100<br>100<br>100<br><br>N/A | 0.03 – 800<br>0.03 – 800<br>0.03 – 800<br>0.03 – 800<br><br>N/A | 50 | ±3000 µV<br>±3000 µV<br>±3000 µV<br>±3000 µV<br><br>N/A |
| DREAMS | University of MONS - TCTS Laboratory, and Université Libre de Bruxelles - CHU de Charleroi Sleep Laboratory, Belgium | Brainnet (MEDATEC, Belgium) | EEG1 Cz-A1<br>EEG2 Fp1-A2<br>EOG (P8-P18)<br>EMG submental<br><br>ECG modified lead-II | 200<br>200<br>200<br>200<br><br>200 | 0.16 – 70<br>0.16 – 70<br>0.16 – 70<br>10 – 70<br><br>0.16 - 70 | 50 | ±800 µV<br>±800 µV<br>±800 µV<br>±800 µV<br><br>±3 mV |
| ISRUC | Sleep Medicine Centre of the Hospital of Coimbra University (CHUC), Portugal | SomnoStar Pro (SensorMedics Corporation, USA) | EEG1 C3-M2<br>EEG2 C4-M1<br>EOG E2-E1<br>EMG submental<br><br>ECG | 200<br>200<br>200<br>200<br><br>200 | 0.3 – 35*<br>0.3 – 35*<br>0.3 – 35*<br>10 – 70*<br><br>N/A<br><br>*All Notch filtered 50 Hz as well | 50 | ±25 µV<br>±25 µV<br>±25 µV<br>±101 µV<br><br>±87 µV |

*Supplementary Table 1. Summary characteristics of the datasets included in the experimentation (continuation)*

| Dataset | Population characteristics | Reference for sleep scoring | Number scorers | Number instances | Dataset class distribution (Proportion of W, N1, N2, N3, R) |
|---|---|---|---|---|---|
| HMC | Random selection of 159 PSG recordings from the sleep center database containing a mix of patients affected of different sleep disorders. Selection includes both in-hospital and ambulatory recordings | AASM 2.4 2017 | 10 | 139145 | 0.17, 0.11, 0.36, 0.19, 0.15 |
| Dublin | 25 (21M, 4F) full overnight PSGs from adult subjects with suspected sleep-disordered breathing (possible diagnosis of obstructive sleep apnea, central sleep apnea or primary snoring). Subjects had to be above 18 years of age, with no known cardiac disease, autonomic dysfunction, and not on medication known to interfere with heart rate. Age: $50 \pm 10$ years, range 28-68 years; BMI: $31.6 \pm 4.0$ kg/m², range 25.1-42.5 kg/m²; AHI: $24.1 \pm 20.3$, range 1.7-90.9 | Rechtschaffen and Kales | 1 | 20774 | 0.23, 0.16, 0.34, 0.13, 0.15 |
| SHHS | Random subset of 100 PSG recordings gathered from the Sleep Health Heart Study (SHHS) follow-up 2. Inclusion criteria included age 40 years or older, no history of treatment of sleep apnea, no tracheostomy, and no current home oxygen therapy. Sample does not discard patients with cardiovascular disorders. | Modified Rechtschaffen and Kales (check manual of operations for details [2]) | 2-5 | 108965 | 0.25, 0.04, 0.44, 0.11, 0.16 |
| Telemetry | 44 whole-night PSGs obtained from a study of temazepam effects on sleep in 22 Caucasian males and females without other medication. Subjects had mild difficulty falling asleep but were otherwise healthy. The PSGs of about 9 hours were recorded in the hospital during two nights, one of which was after temazepam intake, and the other of which was after placebo intake | Rechtschaffen and Kales | 8 | 42691 | 0.10, 0.09, 0.47, 0.15, 0.20 |
| DREAMS | 20 whole-night PSG recordings coming from healthy subjects. These recordings were specifically selected for their clarity (i.e. that they contain few artifacts) and come from persons, free of any medication, volunteers in other research projects, conducted in the sleep lab | AASM 2007 | 1 | 20242 | 0.18, 0.07, 0.41, 0.19, 0.15 |
| ISRUC | 100 subjects (55 male, 45 female) with evidence of having sleep disorders (subgroup-I). Most of the subjects have detected sleep apnea events; the subjects could be under medication, but all were in position to breathe without the help of machine. Age 20-85, avg.=51, std.=16 years | AASM 2007 | 2 (scorings from expert 1 are used in this study as reference) | 90187 | 0.23, 0.13, 0.31, 0.19, 0.13 |